\documentclass[runningheads]{llncs}
\usepackage[T1]{fontenc}
\usepackage{multirow}
\usepackage{makecell}
\usepackage{graphicx}
\usepackage{amsmath}
\usepackage{amssymb}
\usepackage{color}
\usepackage{xfrac}
\usepackage{wrapfig}
\usepackage{cite}
\usepackage{floatrow}
\usepackage{pifont}
\newfloatcommand{capbtabbox}{table}[][\FBwidth]
\usepackage[misc]{ifsym}
\usepackage[colorlinks,
            urlcolor=red,
            hyperfootnotes = True,
            citecolor=green]{hyperref}
\newcommand{\PreserveBackslash}[1]{\let\temp=\\#1\let\\=\temp}
\newcolumntype{C}[1]{>{\PreserveBackslash\centering}p{#1}}
\newcolumntype{R}[1]{>{\PreserveBackslash\raggedleft}p{#1}}
\newcolumntype{L}[1]{>{\PreserveBackslash\raggedright}p{#1}}

\begin{document}
\title{Multi-scale Cross-restoration Framework for Electrocardiogram Anomaly Detection}
\titlerunning{Multi-scale Cross-restoration Framework for ECG Anomaly Detection}

\author{Aofan Jiang\inst{1,2}\thanks{Equal contribution}, Chaoqin Huang\inst{1,2,4\star}, Qing Cao\inst{3}, Shuang Wu\inst{3}, Zi Zeng\inst{3}, \\Kang Chen\inst{3}, Ya Zhang\inst{1,2}, and Yanfeng Wang\inst{1,2}\textsuperscript{(\Letter)}
}
%index{Jiang, Aofan}
%index{Huang, Chaoqin}
%index{Cao, Qing}
%index{Wu, Shuang}
%index{Zeng, Zi}
%index{Chen, Kang}
%index{Zhang, Ya}
%index{Wang, Yanfeng}
\authorrunning{A. Jiang et al.}
\institute{Shanghai Jiao Tong University \and
Shanghai AI Laboratory \and
Ruijin Hospital, Shanghai Jiao Tong University School of Medicine \and
National University of Singapore\\
{\scriptsize \email{\{stillunnamed, huangchaoqin, shuang-renata, zengzidoct, ya\_zhang, wangyanfeng\}@sjtu.edu.cn, \{cq30553, ck11208\}@rjh.com.cn} 
}}

\maketitle

\begin{abstract}
Electrocardiogram (ECG) is a widely used diagnostic tool for detecting heart conditions. Rare cardiac diseases may be underdiagnosed using traditional ECG analysis, considering that no training dataset can exhaust all possible cardiac disorders. This paper proposes using anomaly detection to identify any unhealthy status, with normal ECGs solely for training. However, detecting anomalies in ECG can be challenging due to significant inter-individual differences and anomalies present in both global rhythm and local morphology. To address this challenge, this paper introduces a novel multi-scale cross-restoration framework for ECG anomaly detection and localization that considers both local and global ECG characteristics. The proposed framework employs a two-branch autoencoder to facilitate multi-scale feature learning through a masking and restoration process, with one branch focusing on global features from the entire ECG and the other on local features from heartbeat-level details, mimicking the diagnostic process of cardiologists. Anomalies are identified by their high restoration errors. To evaluate the performance on a large number of individuals, this paper introduces a new challenging benchmark with signal point-level ground truths annotated by experienced cardiologists. The proposed method demonstrates state-of-the-art performance on this benchmark and two other well-known ECG datasets. The benchmark dataset and source code are available at: \url{https://github.com/MediaBrain-SJTU/ECGAD}
\end{abstract}

\section{Introduction}
The electrocardiogram (ECG) is a monitoring tool widely used to evaluate the heart status of patients and provide information on cardiac electrophysiology. Developing automated analysis systems capable of detecting and identifying abnormal signals is crucial in light of the importance of ECGs in medical diagnosis and the need to ease the workload of clinicians. However, training a classifier on labeled ECGs that focus on specific diseases may not recognize new abnormal statuses that were not encountered during training, given the diversity and rarity of cardiac diseases~\cite{kiranyaz2015real, wang2021automated, shaker2020generalization}. On the other hand, anomaly detection, which is trained only on normal healthy data, can identify any potential abnormal status and avoid the failure to detect rare cardiac diseases~\cite{li2020survey, venkatesan2018ecg, shen2021time}.

The current anomaly detection techniques, including one-class discriminative approaches~\cite{chalapathy2018ocad1, ruff2018ocad2}, reconstruction-based approaches~\cite{zong2018deep, schlegl2019f}, and self-supervised learning-based approaches~\cite{chen2019self, ye2020attribute}, all operate under the assumption that models trained solely on normal data will struggle to process anomalous data and thus the substantial drop in performance presents an indication of anomalies. While anomaly detection has been widely used in the medical field to analyze medical images~\cite{mao2020uncertainty, wolleb2022diffusion} and time-series data~\cite{tranad, zheng2022task}, detecting anomalies in ECG data is particularly challenging due to the substantial inter-individual differences and the presence of anomalies in both global rhythm and local morphology. So far, few studies have investigated anomaly detection in ECG ~\cite{liu2022time,zheng2022task}. TSL~\cite{zheng2022task} uses expert knowledge-guided amplitude- and frequency-based data transformations to simulate anomalies for different individuals. BeatGAN~\cite{liu2022time} employs a generative adversarial network to separately reconstruct normalized heartbeats instead of the entire raw ECG signal. While BeatGAN alleviates individual differences, it neglects the important global rhythm information of the ECG.

This paper proposes a novel multi-scale cross-restoration framework for ECG anomaly detection and localization. To our best knowledge, this is the first work to integrate both local and global characteristics for ECG anomaly detection. To take into account multi-scale data, the framework adopts a two-branch autoencoder architecture, with one branch focusing on global features from the entire ECG and the other on local features from heartbeat-level details. A multi-scale cross-attention module is introduced, which learns to combine the two feature types for making the final prediction. This module imitates the diagnostic process followed by experienced cardiologists who carefully examine both the entire ECG and individual heartbeats to detect abnormalities in both the overall rhythm and the specific local morphology of the signal~\cite{ecgbook}. Each of the branches employs a masking and restoration strategy, \emph{i.e.,} the model learns how to perform temporal-dependent signal inpainting from the adjacent unmasked regions within a specific individual. Such context-aware restoration has the advantage of making the restoration less susceptible to individual differences. During testing, anomalies are identified as samples or regions with high restoration errors.

To comprehensively evaluate the performance of the proposed method on a large number of individuals, we adopt the public PTB-XL database~\cite{wagner2020ptb} with only patient-level diagnoses, and ask experienced cardiologists to provide signal point-level localization annotations. The resulting dataset is then introduced as a large-scale challenging benchmark for ECG anomaly detection and localization. The proposed method is evaluated on this challenging benchmark as well as on two traditional ECG anomaly detection benchmarks~\cite{moody2001impact, keogh2004hot}. The experimental results have shown that the proposed method outperforms several state-of-the-art methods for both anomaly detection and localization, highlighting its potential for real-world clinical diagnosis.

\section{Method}
In this paper, we focus on unsupervised anomaly detection and localization on ECGs, training based on only normal ECG data. Formally, given a set of $N$ normal ECGs denoted as $\{x_i,i=1,...,N\}$, where $x_i \in \mathbb{R}^D$ represents the vectorized representation of the $i$-th ECG consisting of $D$ signal points, the objective is to train a computational model capable of identifying whether a new ECG is normal or anomalous, and localize the regions of anomalies in abnormal ECGs.

\begin{figure}[t]
\centering
\includegraphics[width=1.0\textwidth]{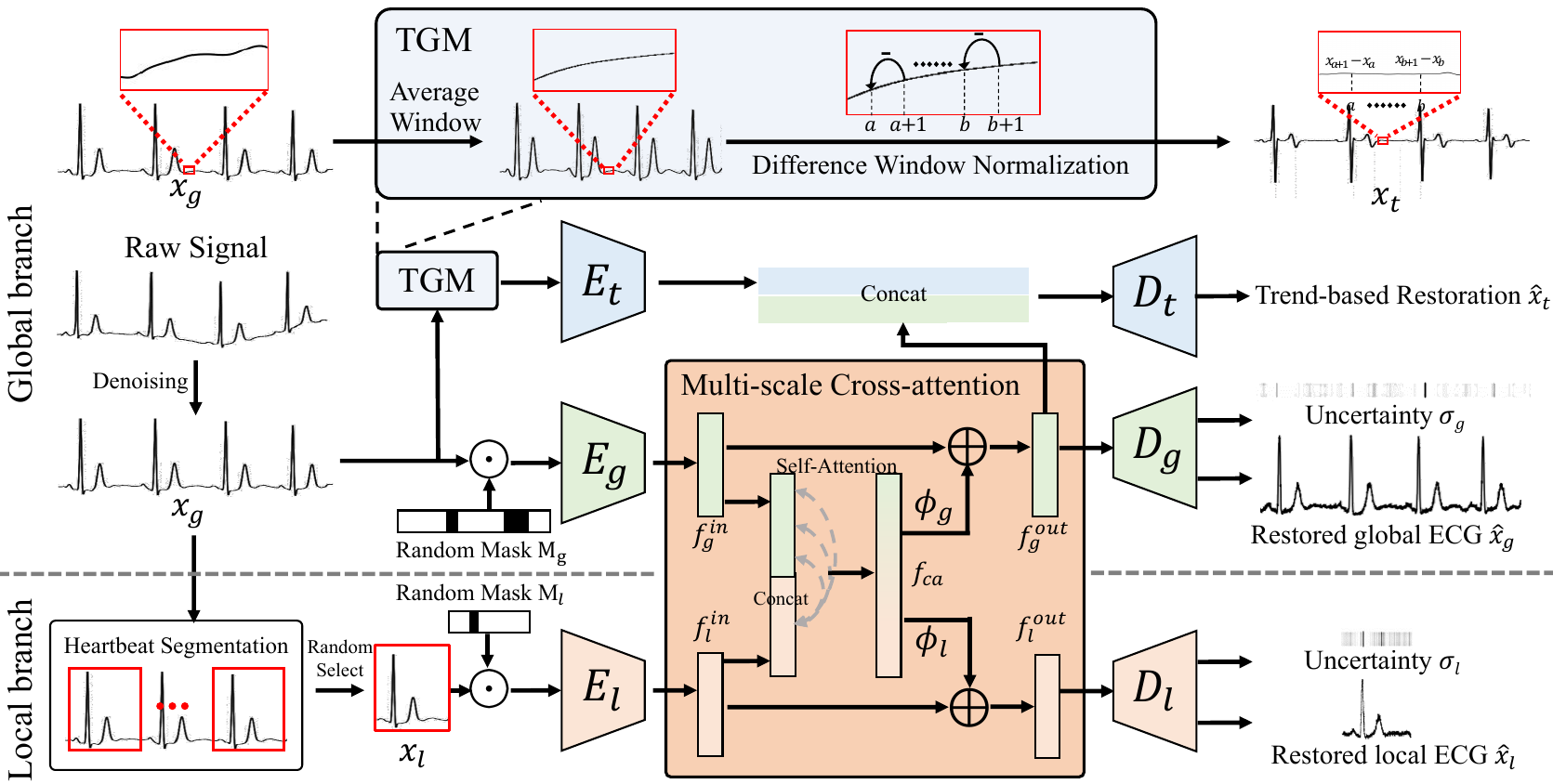}
\caption{The multi-scale cross-restoration framework for ECG anomaly detection.}
\label{fig:model}
\end{figure}

\subsection{Multi-scale Cross-restoration}
In Fig.~\ref{fig:model}, we present an overview of our two-branch framework for ECG anomaly detection. One branch is responsible for learning global ECG features, while the other focuses on local heartbeat details. Our framework comprises four main components: (i) masking and encoding, (ii) multi-scale cross-attention module, (iii) uncertainty-aware restoration, and (iv) trend generation module. We provide detailed explanations of each of these components in the following sections.

\noindent\textbf{Masking and Encoding.} 
Given a pair consisting of a global ECG signal $x_{g}\in \mathbb{R}^D$ and a randomly selected local heartbeat $x_{l}\in\mathbb{R}^d$ segmented from $x_{g}$ for training, as shown in Fig.~\ref{fig:model}, we apply two random masks, $M_g$ and $M_l$, to mask $x_{g}$ and $x_{l}$, respectively. To enable multi-scale feature learning, $M_l$ is applied to a consecutive small region to facilitate detail restoration, while $M_g$ is applied to several distinct regions distributed throughout the whole sequence to facilitate global rhythm restoration. The masked signals are processed separately by global and local encoders, $E_g$ and $E_l$, resulting in global feature $f^{in}_g = E_g (x_{g}\odot M_g)$ and local feature $f^{in}_l = E_l (x_{l}\odot M_l)$, where $\odot$ denotes the element-wise product.

\noindent\textbf{Multi-scale Cross-attention.}
To capture the relationship between global and local features, we use the self-attention mechanism~\cite{vaswani2017attention} on the concatenated feature of $f^{in}_g$ and $f^{in}_l$. Specifically, the attention mechanism is expressed as $Attention(Q,K,V)=\mathrm{softmax}(\frac{QK^T}{\sqrt{d_k}})V$, where $Q, K, V$ are identical input terms, while $\sqrt{d_k}$ is the square root of the feature dimension used as a scaling factor. Self-attention is achieved by setting $Q=K=V=concat(f^{in}_g,f^{in}_l)$. The cross-attention feature, $f_{ca}$, is obtained from the self-attention mechanism, which dynamically weighs the importance of each element in the combined feature. To obtain the final outputs of the global and local features, $f^{out}_g$ and $f^{out}_l$, containing cross-scale information, we consider residual connections: $f^{out}_g = f^{in}_g + \phi_g(f_{ca}),\ f^{out}_l = f^{in}_l + \phi_l(f_{ca})$, where $\phi_g (\cdot)$ and $\phi_l (\cdot)$ are MLP architectures with two fully connected layers.

\noindent\textbf{Uncertainty-aware Restoration.}
Targeting signal restorations, features of $f^{out}_g$ and $f^{out}_l$ are decoded by two decoders, $D_g$ and $D_l$, to obtain restored signals $\hat{x}_g$ and $\hat{x}_l$, respectively, along with corresponding restoration uncertainty maps $\sigma_g$ and $\sigma_l$ measuring the difficulty of restoration for various signal points, where $\hat{x}_g, \sigma_g = D_g(f^{out}_g),\ \hat{x}_l, \sigma_l = D_l(f^{out}_l)$. An uncertainty-aware restoration loss is used to incorporate restoration uncertainty into the loss functions,
\begin{equation}
\mathcal{L}_{global}=\sum_{k=1}^D \{\frac{(x_{g}^{k}-\hat{x}_{g}^{k})^2}{\sigma^{k}_{g}}+\log \sigma^{k}_{g}\}, \ \ 
\mathcal{L}_{local}=\sum_{k=1}^d \{\frac{(x_{l}^{k}-\hat{x}_{l}^{k})^2}{\sigma^{k}_l}+\log \sigma^{k}_{l}\},
\end{equation}
where for each function, the first term is normalized by the corresponding uncertainty, and the second term prevents predicting a large uncertainty for all restoration pixels following~\cite{mao2020uncertainty}. The superscript $k$ represents the position of the $k$-th element of the signal. It is worth noting that, unlike~\cite{mao2020uncertainty}, the uncertainty-aware loss is used for restoration, but not for reconstruction.

\noindent\textbf{Trend Generation Module.}
The trend generation module (TGM) illustrated in Fig.~\ref{fig:model} generates a smooth time-series trend $x_t \in \mathbb{R}^D$ by removing signal details, which is represented as the smooth difference between adjacent time-series signal points. An autoencoder ($E_t$ and $D_t$) encodes the trend information into $E_t(x_t)$, which are concatenated with the global feature $f^{out}_g$ to restore the global ECG $\hat{x}_{t} = D_t(\text{concat}(E_t(x_{t}),f^{out}_g))$. The restoration loss is defined as the Euclidean distance between $x_{g}$ and $\hat{x}_{t}$, $\mathcal{L}_{trend}=\sum_{k=1}^D (x_{g}^{k}-\hat{x}_{t}^{k})^2.$ This process guides global feature learning using time-series trend information, emphasizing rhythm characteristics while de-emphasizing morphological details.

\noindent\textbf{Loss Function.}
The final loss function for optimizing our model during the training process can be written as
\begin{equation}
\mathcal{L} = \mathcal{L}_{global} + \alpha \mathcal{L}_{local}+\beta \mathcal{L}_{trend}, 
\end{equation}
where $\alpha$ and $\beta$ are trade-off parameters weighting the loss function. For simplicity, we adopt $\alpha = \beta = 1.0$ as the default.

\subsection{Anomaly Score Measurement} 
For each test sample $x$, local ECGs from the segmented heartbeat set $\{x_{l,m},m=1,...,M\}$ are paired with the global ECG $x_{g}$ one at a time as inputs. The anomaly score $\mathcal{A}(x)$ is calculated to estimate the abnormality, 
\begin{equation}\label{eq:ad}
\mathcal{A}(x)= \sum_{k=1}^D \frac{(x_{g}^{k}-\hat{x}_g^{k})^2}{\sigma^{k}_g} + \sum_{m=1}^M \sum_{k=1}^{d} \frac{(x^{k}_{l,m}-\hat{x}_{l,m}^{k})^2}{\sigma^{k}_{l,m}} + \sum_{k=1}^D (x_{g}^{k}-\hat{x}_{t}^{k})^2,
\end{equation}
where the three terms correspond to global restoration, local restoration, and trend restoration, respectively. For localization, an anomaly score map is generated in the same way as Eq.~\eqref{eq:ad}, but without summing over the signal points. The anomalies are indicated by relatively large anomaly scores, and vice versa.

\section{Experiments}
\textbf{Datasets}
Three publicly available ECG datasets are used to evaluate the proposed method, including PTB-XL~\cite{wagner2020ptb}, MIT-BIH~\cite{moody2001impact}, and Keogh ECG~\cite{keogh2004hot}. 
\begin{itemize} 
\item \textbf{PTB-XL} database includes clinical 12-lead ECGs that are 10 seconds in length for each patient, with only \underline{patient-level} annotations. To build a new challenging anomaly detection and localization \textbf{benchmark}, 8167 normal ECGs are used for training, while 912 normal and 1248 abnormal ECGs are used for testing. We provide \underline{signal point-level} annotations of 400 ECGs, including 22 different abnormal types, that were annotated by two experienced cardiologists. To our best knowledge, we are the first to explore ECG anomaly detection and localization across various patients on such a complex and large-scale database. 
\item \textbf{MIT-BIH} arrhythmia dataset divides the ECGs from 44 patients into independent heartbeats based on the annotated R-peak position, following~\cite{liu2022time}. 62436 normal heartbeats are used for training, while 17343 normal and 9764 abnormal beats are used for testing, with \underline{heartbeat-level} annotations. 
\item \textbf{Keogh ECG} dataset includes 7 ECGs from independent patients, evaluating anomaly localization with \underline{signal point-level} annotations. For each ECG, there is an anomaly subsequence that corresponds to a pre-ventricular contraction, while the remaining sequence is used as normal data to train the model. The ECGs are partitioned into fixed-length sequences of 320 by a sliding window with a stride of 40 during training and 160 during testing.    
\end{itemize}

\noindent\textbf{Evaluation Protocols}
The performance of anomaly detection and localization is quantified using the area under the Receiver Operating Characteristic curve (AUC), with a higher AUC value indicating a better method. To ensure comparability across different annotation levels, we used patient-level, heartbeat-level, and signal point-level AUC for each respective setting. For heartbeat-level classification, the F1 score is also reported following~\cite{liu2022time}.

\noindent\textbf{Implementation Details}
The ECG is pre-processed by a Butterworth filter and Notch filter~\cite{van2019heartpy} to remove high-frequency noise and eliminate ECG baseline wander. The R-peaks are detected with an adaptive threshold following~\cite{van2019rpeak}, which does not require any learnable parameters. The positions of the detected R-peaks are then used to segment the ECG sequence into a set of heartbeats.

We use a convolutional-based autoencoder, following the architecture proposed in~\cite{liu2022time}. The model is trained using the AdamW optimizer with an initial learning rate of 1e-4 and a weight decay coefficient of 1e-5 for 50 epochs on a single NVIDIA GTX 3090 GPU, with a single cycle of cosine learning rate used for decay scheduling. The batch size is set to 32. During testing, the model requires 2365M GPU memory and achieves an inference speed of 4.2 fps.

\begin{table*}[t]
\centering
\begin{floatrow}
\capbtabbox{\setlength{\tabcolsep}{0.95pt}{
\begin{tabular}{C{1.8cm}C{0.9cm}|C{1.4cm}C{1.5cm}}
\hline
Method & Year & detection & localization \\
\hline
DAGMM~\cite{zong2018deep} & 2018 & 0.782 & 0.688\\
MADGAN~\cite{li2019mad} & 2019 & 0.775 & 0.708\\
USAD~\cite{audibert2020usad} & 2020 & 0.785 & 0.683\\
TranAD~\cite{tranad}  & 2022 & 0.788 & 0.685\\
AnoTran~\cite{xu2022anomaly} & 2022 & 0.762 & 0.641 \\
TSL~\cite{zheng2022task} & 2022 & 0.757 & 0.509\\
BeatGAN~\cite{liu2022time} & 2022 & \underline{0.799} & \underline{0.715}\\
\hline
Ours & 2023 & \textbf{0.860} & \textbf{0.747} \\
\hline
\end{tabular}}
}{
 \caption{Anomaly detection and anomaly localization results on PTB-XL database. Results are shown in the patient-level AUC for anomaly detection and the signal point-level AUC for anomaly localization, respectively. The best-performing method is in \textbf{bold}, and the second-best is \underline{underlined}.}
 \label{tal:ptxb}
}
\capbtabbox{\setlength{\tabcolsep}{0.95pt}{
\begin{tabular}{C{1.9cm}C{0.9cm}|C{1.0cm}C{1.0cm}} 
\hline
Method & Year & F1 & AUC \\
\hline
DAGMM~\cite{zong2018deep} & 2018 & 0.677 & 0.700\\
MSCRED~\cite{zhang2019deep} & 2019 & 0.778 & 0.627\\
USAD~\cite{audibert2020usad} & 2020 & 0.384 & 0.352\\
TranAD~\cite{tranad} & 2022 & 0.621 & 0.742 \\
AnoTran~\cite{xu2022anomaly} & 2022 & 0.650 & 0.770 \\
TSL~\cite{zheng2022task}  & 2022 & 0.750 & 0.894\\
BeatGAN~\cite{liu2022time} & 2022 & \underline{0.816} & \underline{0.945}\\
\hline
Ours & 2023 & \textbf{0.883} & \textbf{0.969} \\
\hline
\end{tabular}
}
}{
 \caption{Anomaly detection results on MIT-BIH dataset, comparing with state-of-the-arts. Results are shown in terms of the AUC and F1 score for heartbeat-level classification. The best-performing method is in \textbf{bold}, and the second-best is \underline{underlined}.}
 \label{tal:mit}
}
\end{floatrow}
\end{table*}

\begin{table}[t]
 \caption{Anomaly localization results on Keogh ECG~\cite{keogh2004hot} dataset, comparing with several state-of-the-arts. Results are shown in the signal point-level AUC. The best-performing method is in \textbf{bold}, and the second-best is \underline{underlined}.
 }
\centering
\begin{tabular}{C{2.0cm}C{1.0cm}|C{1.0cm}C{1.0cm}C{1.0cm}C{1.0cm}C{1.0cm}C{1.0cm}C{1.0cm}|C{1.0cm}} 
\hline
Methods & Year & A & B & C & D & E & F & G & Avg \\
\hline
DAGMM~\cite{zong2018deep} & 2018     & 0.672         & 0.612                & 0.805           &  0.713         & 0.457    & 0.662 & 0.676           & 0.657 \\
MSCRED~\cite{zhang2019deep} & 2019 & 0.667 & 0.633 & 0.798 & 0.714 & 0.461 & 0.746 & 0.659 & 0.668\\
MADGAN~\cite{li2019mad} & 2019                             & 0.688         & \textbf{0.702}               & \textbf{0.833}         &   0.664         & 0.463     & 0.692   & 0.678          & 0.674 \\
USAD~\cite{audibert2020usad}  & 2020                              & 0.667          & 0.616                & 0.795           &  0.715        & 0.462     & 0.649 & 0.680          & 0.655 \\
GDN ~\cite{deng2021graph}& 2021 & 0.695 & 0.611 & 0.790 & 0.674 & 0.458 & 0.648 & 0.671 & 0.650\\
CAE-M~\cite{zhang2021unsupervised} & 2021 & 0.657 & 0.618 & 0.802 & 0.715 & 0.457 & 0.708 & 0.671 & 0.661\\
TranAD~\cite{tranad}  & 2022                        & 0.647 & 0.623        & \underline{0.820}  &   0.720         & 0.446   & \textbf{0.780} & 0.680             & 0.674  \\
AnoTran~\cite{xu2022anomaly}  & 2022                         & 0.739        & 0.502               & 0.792          &   \underline{0.799} & 0.498   &0.748 &0.711   & 0.684 \\
BeatGAN~\cite{liu2022time}  & 2022                             & \underline{0.803}         & 0.623                 & 0.783           &  0.747         & \underline{0.506}    & 0.757 & \textbf{0.852}          & \underline{0.724} \\
\hline
Ours   & 2023 & \textbf{0.832} & \underline{0.641}        & 0.819 &   \textbf{0.815} & \textbf{0.543} & \underline{0.760} & \underline{0.833}     & \textbf{0.749}\\
\hline
\end{tabular}
\label{tab:keo}
\end{table}

\subsection{Comparisons with State-of-the-Arts}
We compare our method with several time-series anomaly detection methods, including heartbeat-level detection method BeatGAN~\cite{liu2022time}, patient-level detection method TSL~\cite{zheng2022task}, and several signal point-level anomaly localization methods~\cite{zong2018deep,zhang2019deep,li2019mad,audibert2020usad,deng2021graph,zhang2021unsupervised,xu2022anomaly,tranad}. For a fair comparison, we re-trained all the methods under the same experimental setup. For those methods originally designed for signal point-level tasks only~\cite{zong2018deep,li2019mad,audibert2020usad,xu2022anomaly,tranad}, we use the mean value of anomaly localization results as their heartbeat-level or patient-level anomaly scores.

\begin{figure}[t]
\centering
\includegraphics[width=0.95\textwidth]{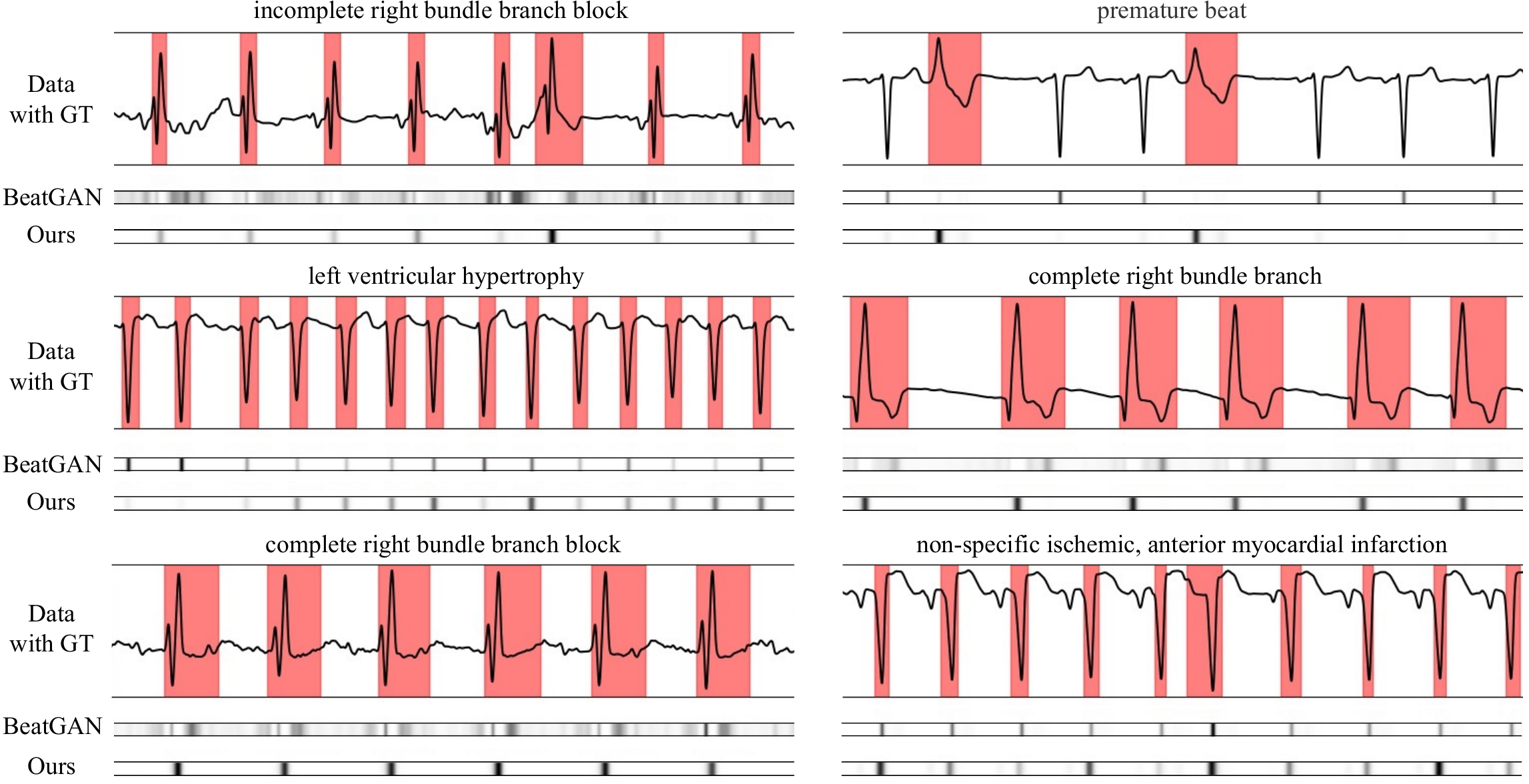}
\caption{Anomaly localization visualization on PTB-XL with different abnormal types. Ground truths are highlighted in red boxes on the ECG data, and anomaly localization results for each case, compared with the state-of-the-art method, are attached below.}
\label{fig:vis}
\end{figure}

\noindent\textbf{Anomaly Detection} The anomaly detection performance on PTB-XL is summarized in Table~\ref{tal:ptxb}. The proposed method achieves 86.0\% AUC in patient-level anomaly detection and outperforms all baselines by a large margin (10.3\%). Table~\ref{tal:mit} displays the comparison results on MIT-BIH, where the proposed method achieves a heartbeat-level AUC of 96.9\%, showing an improvement of 2.4\% over the state-of-the-art BeatGAN (94.5\%). Furthermore, the F1-score of the proposed method is 88.3\%, which is 6.7\% higher than BeatGAN (81.6\%).

\noindent\textbf{Anomaly Localization} Table~\ref{tal:ptxb} presents the results of anomaly localization on our proposed benchmark for multiple individuals. The proposed method achieves a signal point-level AUC of 74.7\%, outperforming all baselines (3.2\% higher than BeatGAN). It is worth noting that TSL, which is not designed for localization, shows poor performance in this task. Table~\ref{tab:keo} shows the signal point-level anomaly localization results for each independent individual on Keogh ECG. Overall, the proposed method achieves the best or second-best performance compared to other methods on six subsets and the highest mean AUC among all subsets (74.9\%, 2.5\% higher than BeatGAN), indicating its effectiveness. The proposed method shows a lower standard deviation ($\pm10.5$) across the seven subsets compared to TranAD ($\pm11.3$) and BeatGAN ($\pm11.0$), which indicates good generalizability of the proposed method across different subsets.

\noindent\textbf{Anomaly Localization Visualization}
We present visualization results of anomaly localization on several samples from our proposed benchmark in Fig.~\ref{fig:vis}, with ground truths annotated by experienced cardiologists. Regions with higher anomaly scores are indicated by darker colors. Our proposed method outperforms BeatGAN in accurately localizing various types of ECG anomalies, including both periodic and episodic anomalies, such as incomplete right bundle branch block and premature beats. Our method though provides narrower localization results than ground truths, as it is highly sensitive to abrupt unusual changes in signal values, but still represents the important areas for anomaly identification, a fact confirmed by experienced cardiologists.

\begin{table}[t]
\centering
\begin{floatrow}
\capbtabbox{\setlength{\tabcolsep}{2.0pt}{
\begin{tabular}{C{1.0cm}C{1.0cm}C{1.0cm}C{1.0cm}|C{2.0cm}} 
\hline
MR & MC & UL & TGM & AUC\\
\hline
&  &  &  & 70.4$_{\pm 0.3}$ \\
\hline
\checkmark &  &  &  &  80.4$_{\pm 0.7}$ \\
& \checkmark &  &  & 80.3$_{\pm 0.3}$ \\
&  & \checkmark &  & 72.8$_{\pm 2.0}$ \\
&  &  & \checkmark & 71.2$_{\pm 0.5}$ \\
\hline
\checkmark & \checkmark &  &  & 84.8$_{\pm 0.8}$ \\
\checkmark & \checkmark & \checkmark &  & 85.2$_{\pm 0.4}$ \\
\checkmark & \checkmark & \checkmark & \checkmark & \textbf{86.0}$_{\pm 0.1}$ \\
\hline
\end{tabular}}
}{
\caption{Ablation studies on PTB-XL dataset. Factors under analysis are: the masking and restoring (MR), the multi-scale cross-attention (MC), the uncertainty loss function (UL), and the trend generation module (TGM). Results are shown in the patient-level AUC in \% of five runs. The best-performing method is in \textbf{bold}.}
\label{tab:abl}
}
\capbtabbox{\setlength{\tabcolsep}{2.0pt}{
\begin{tabular}{C{1.8cm}|C{2.0cm}}
\hline
Mask Ratio & AUC \\
\hline
0\% & 80.2$_{\pm 0.2}$\\
10\% & 85.2$_{\pm 0.2}$\\
20\% & \underline{85.5}$_{\pm 0.3}$\\
30\% & \textbf{86.0}$_{\pm 0.1}$\\
40\% & 84.9$_{\pm 0.3}$\\
50\% & 83.8$_{\pm 0.1}$\\
60\% & 82.9$_{\pm 0.1}$\\
70\% & 75.8$_{\pm 1.0}$\\
\hline
\end{tabular}}
}{
 \caption{Sensitivity analysis w.r.t. mask ratio on PTB-XL dataset. Results are shown in the patient-level AUC of five runs. The best-performing method is in \textbf{bold}, and the second-best is \underline{underlined}.}
 \label{tal:sen}
 \small
}
\end{floatrow}
\end{table}

\subsection{Ablation Study and Sensitivity Analysis}
Ablation studies were conducted on PTB-XL to confirm the effectiveness of individual components of the proposed method. Table~\ref{tab:abl} shows that each module contributes positively to the overall performance of the framework. When none of the modules were employed, the method becomes a ECG reconstruction approach with a naive L2 loss and lacks cross-attention in multi-scale data. When individually adding the MR, MC, UL, and TGM modules to the baseline model without any of them, the AUC values improve from 70.4\% to 80.4\%, 80.3\%, 72.8\%, and 71.2\%, respectively, demonstrating the effectiveness of each module. Moreover, as the modules are added in sequence, the performance improves step by step from 70.4\% to 86.0\% in AUC, highlighting the combined impact of all modules on the proposed framework.

We conduct a sensitivity analysis on the mask ratio, as shown in Table~\ref{tal:sen}. Restoration with a 0\% masking ratio can be regarded as reconstruction, which takes an entire sample as input and its target is to output the input sample. Results indicate that the model's performance first improves and then declines as the mask ratio increases from 0\% to 70\%. This trend is due to the fact that a low mask ratio can limit the model's feature learning ability during restoration, while a high ratio can make it increasingly difficult to restore the masked regions. Therefore, there is a trade-off between maximizing the model's potential and ensuring a reasonable restoration difficulty. The optimal mask ratio is 30\%, which achieves the highest anomaly detection result (86.0\% in AUC).

\section{Conclusion}
This paper proposes a novel framework for ECG anomaly detection, where features of the entire ECG and local heartbeats are combined with a masking-restoration process to detect anomalies, simulating the diagnostic process of cardiologists. A challenging benchmark, with signal point-level annotations provided by experienced cardiologists, is proposed, facilitating future research in ECG anomaly localization. The proposed method outperforms state-of-the-art methods, highlighting its potential in real-world clinical diagnosis.

\subsubsection{Acknowledgement}
This work is supported by the National Key R\&D Program of China (No. 2022ZD0160702), STCSM (No. 22511106101, No. 18DZ2270700, No. 21DZ1100100), 111 plan (No. BP0719010), the Youth Science Fund of National Natural Science Foundation of China (No.7210040772) and National Facility for Translational Medicine (Shanghai) (No.TMSK-2021-501), and State Key Laboratory of UHD Video and Audio Production and Presentation.

\bibliographystyle{splncs04}
\bibliography{mybib}

\newpage

\section{Appendix}

\begin{table}[!h]
\caption{Comparisons between ECG anomaly detection benchmarks.}
\centering
\scalebox{0.9}{
%\resizebox{\textwidth}{!}{
\begin{tabular}{C{1.5cm}|C{1.0cm}C{1.0cm}C{2.0cm}C{1.6cm}C{2.3cm}C{1.2cm}C{1.8cm}} 
\hline
\multirow{3}{*}{Datasets} & \multicolumn{2}{c}{\#Patients} & \multicolumn{3}{c}{Annotation Types} & \multirow{3}{*}{\#Leads} & \multirow{3}{*}{\makecell{\#Abnormal\\Types}}\\ 
& train & test & \makecell{patient\\(detection)} & \makecell{heartbeat\\(detection)} & \makecell{signal point\\(localization)} & & \\
\hline
MIT-BIH & 44 & 44 & all abnormal & \checkmark & \ding{55} & 1 & 1\\
\hline
Keogh & 7 & 7 & all abnormal & - & \checkmark & 2 & 5\\
\hline
\multirow{2}{*}{PTB-XL} & \multirow{2}{*}{8167} & \multirow{2}{*}{2160} & \multirow{2}{*}{\checkmark} & \multirow{2}{*}{-} & \checkmark & \multirow{2}{*}{12} & \multirow{2}{*}{22}\\
&&&&&(our proposed)\\
\hline
\end{tabular}}
\label{tab:dataset}
\end{table}
\vspace{-30pt}

\begin{table}[!h]
 \caption{Abnormal types of ECG anomaly detection benchmarks.}
\centering
\scalebox{0.9}{
\begin{tabular}{C{1.5cm}|C{11.5cm}} 
\hline
Datasets & Abnormal Types\\ 
\hline
MIT-BIH & arrhythmia\\
\hline
Keogh & congestive heart failure, left ventricular hypertrophy, ST elevation, pre-ventricular contraction, ST depression\\
\hline
PTB-XL & left anterior/left posterior fascicular block, incomplete right bundle branch block, incomplete left bundle branch block, complete left bundle branch block, complete right bundle branch block, AV block, non-specific intraventricular conduction disturbance (block), Wolff-Parkinson-White syndrome, left ventricular hypertrophy, right ventricular hypertrophy, left atrial overload/enlargement, right atrial overload/enlargement, septal hypertrophy, anterior myocardial infarction, inferior myocardial infarction, lateral myocardial infarction, posterior myocardial infarction, ischemic in anterior leads, ischemic in inferior leads, non-specific ischemic, ST-T changes, non-specific ST changes\\
\hline
\end{tabular}}
\end{table}

\begin{table}[!h]
\caption{Anomaly detection and localization results on our proposed ECG benchmark. We re-implement the results with the aligned backbone$^*$ for fair comparisons.}
\centering
\scalebox{1.0}{
\begin{tabular}{C{1.7cm}|C{1.1cm}C{1.1cm}C{1.1cm}C{2.0cm}C{1.5cm}C{1.5cm}} 
\hline
\multirow{2}{*}{Methods} & \multirow{2}{*}{Year} & \multicolumn{2}{c}{Multi-Scale} & \multirow{2}{*}{Architecture} & \multicolumn{2}{c}{AUC}\\ 
& & global & local & & detection & localization \\
\hline
DAGMM & 2018 & \checkmark & \ding{55} & \makecell{MLP\\CNN$^*$} & \makecell{0.607\\0.782} & \makecell{0.582\\0.688}\\
\hline
MADGAN & 2019 & \checkmark & \ding{55} & \makecell{MLP\\CNN$^*$} & \makecell{0.607\\0.775} & \makecell{0.599\\0.708}\\
\hline
USAD & 2020 & \checkmark & \ding{55} & \makecell{MLP\\CNN$^*$} & \makecell{0.625\\0.785} & \makecell{0.602\\0.683}\\
\hline
TranAD & 2022 & \checkmark & \ding{55} & Transformer & 0.788 & 0.685\\
\hline
AnoTran & 2022 & \checkmark & \ding{55} & Transformer & 0.762 & 0.641\\
\hline
TSL & 2022 & \checkmark & \ding{55} & CNN & 0.757 & 0.509\\
\hline
BeatGAN & 2022 & \ding{55} & \checkmark & CNN  & 0.799 & 0.715\\
\hline
Ours & 2023 & \checkmark & \checkmark & CNN& 0.860 & 0.747\\
\hline
\end{tabular}}
\end{table}

\begin{table}[!h]
 \caption{Lists of seven patients in Keogh ECG dataset.}
\centering
\scalebox{0.9}{
\begin{tabular}{C{1.5cm}|C{3.0cm}|C{3.0cm}|C{3.0cm}|C{2.7cm}} 
\hline
IDs & A: chfdb\underline{~}chf01\underline{~}275 & B: ltstdb\underline{~}20221\underline{~}43 & C: ltstdb\underline{~}20321\underline{~}240 & D: qtdbsel102\\
\hline
Abnormal Types & congestive heart failure & Left ventricular hypertrophy
& ST elevation & pre-ventricular contraction\\
\hline
IDs & E: stdb\underline{~}308\underline{~}0 & F: chfdb\underline{~}chf13\underline{~}45590 & G: chfdbchf15 \\
\hline
Abnormal Types & ST depression & congestive heart failure & congestive heart failure\\
\hline
\end{tabular}}
\end{table}

\begin{figure}[!h]
\centering
\includegraphics[width=1.0\textwidth]{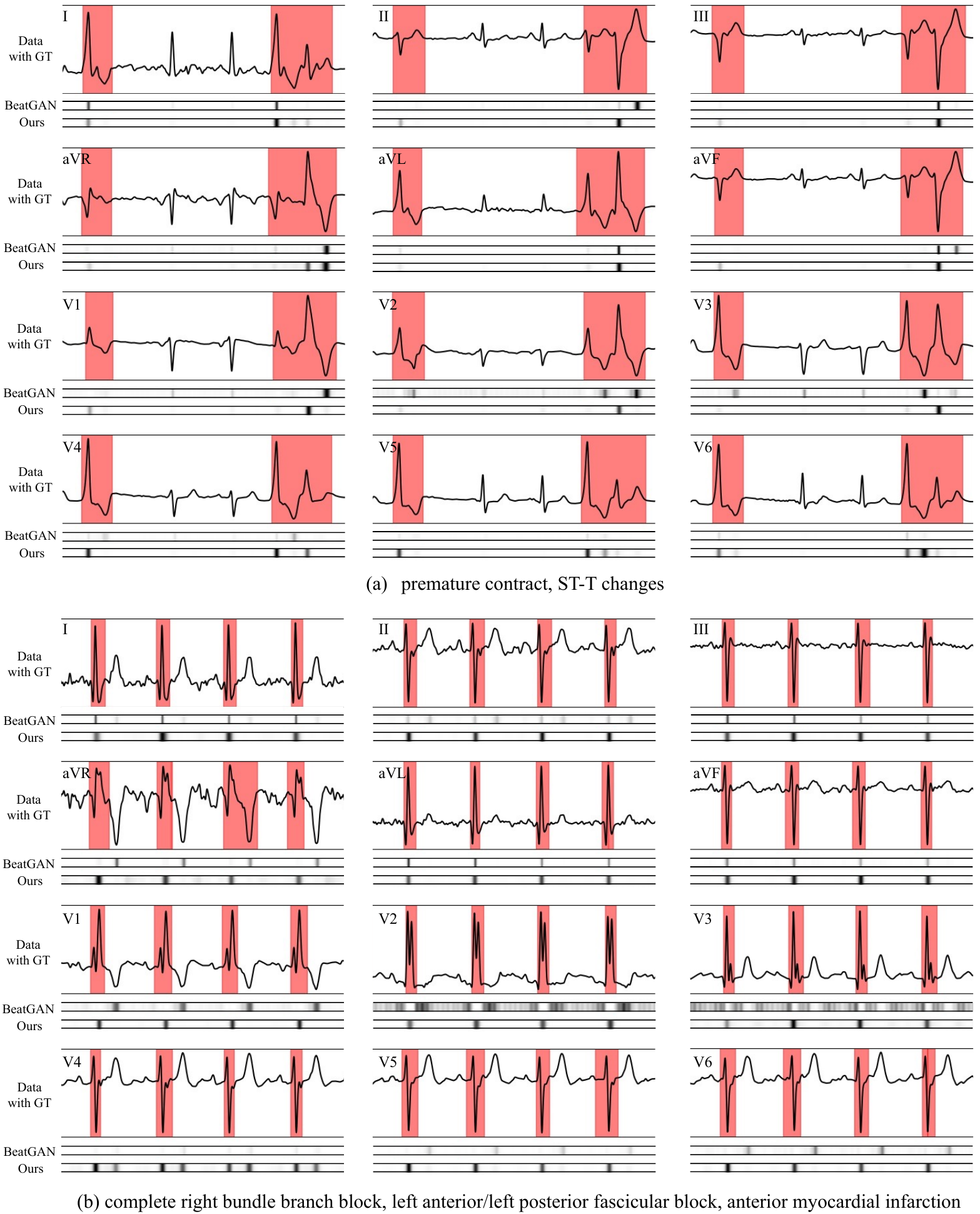}
\caption{Visualization results of two challenging samples on the complete 12-leads ECGs.}
\end{figure}

\end{document}